\documentclass[runningheads，anonymous]{llncs}
\usepackage{comment}
\usepackage{graphicx}
\usepackage{amssymb} 
\usepackage{amsmath} 
\usepackage{multirow}
\usepackage{color}

\begin{document}
\title{RetMIL: Retentive Multiple Instance Learning for Histopathological Whole Slide Image Classification}
%
%
\author{
Hongbo Chu\inst{1} \and
Qiehe Sun\inst{1} \and
Jiawen Li\inst{1} \and
Yuxuan Chen\inst{1} \and
Lizhong Zhang\inst{1} \and
Tian Guan \inst{1} \and
Anjia Han\inst{2} \and
Yonghong He\inst{1}
}

%
%
\institute{Shenzhen International Graduate School, Tsinghua University, China 
\email{\{zhu-hb23,sunqh21,lijiawen21,chenyx23,zhanglz21\}@mails.tsinghua.edu.cn \{heyh, guantian\}@sz.tsinghua.edu.cn}
\and
department of Pathology, The First Affiliated Hospital of Sun Yat-sen University, China \\
\email{hananjia@mail.sysu.edu.cn}}
\maketitle              
\begin{abstract}
Histopathological whole slide image (WSI) analysis with deep learning has become a research focus in computational pathology. The current paradigm is mainly based on multiple instance learning (MIL), in which approaches with Transformer as the backbone are well discussed. These methods convert WSI tasks into sequence tasks by representing patches as tokens in the WSI sequence. However, the feature complexity brought by high heterogeneity and the ultra-long sequences brought by gigapixel size makes Transformer-based MIL suffer from the challenges of high memory consumption, slow inference speed, and lack of performance. To this end, we propose a retentive MIL method called RetMIL, which processes WSI sequences through hierarchical feature propagation structure. At the local level, the WSI sequence is divided into multiple subsequences. Tokens of each subsequence are updated through a parallel linear retention mechanism and aggregated utilizing an attention layer.  At the global level, subsequences are fused into a global sequence, then updated through a serial retention mechanism, and finally the slide-level representation is obtained through a global attention pooling. We conduct experiments on two public CAMELYON and BRACS datasets and an public-internal LUNG dataset, confirming that RetMIL not only achieves state-of-the-art performance but also significantly reduces computational overhead. Our code will be accessed shortly.

\end{abstract}

\section{Introduction}

Pathological slide scanners store microscopic fields of view as the WSI, laying the foundation for automatic diagnostics based on deep learning \cite{madabhushi2009digital}. However, the gigapixel-level resolution and the lack of pixel-level annotations pose significant challenges in developing such intelligent tools. In recent years, with the development of weakly-supervised technologies, MIL methods for WSI analysis have been well studied, which treats WSI as bags and cropped patches as instances. By embedding instances into high-dimensional space for aggregation, slide-level representations can be obtained. MIL methods are generally categorized into instance-level \cite{campanella2019clinical,chikontwe2020multiple,kanavati2020weakly} and embedding-level \cite{abmil,lerousseau2021sparseconvmil11,transmil} approaches. The former has been gradually replaced due to enormous data requirements and weak generalization.

Embedding-level MIL methods generally focus on proposing effective aggregation strategies to obtain more effective WSI representations. Although mean or max pooling is a direct corollary of the MIL theory, dynamically assigning importance scores to patches has proven more effective \cite{abmil,clam}. In addition, due to the wide application of Transformer\cite{transformer}, more research focuses on predicting WSI scores by modeling the correlation between patches through the self-attention mechanism, which helps describe the underlying tumor microenvironment patterns. Transformer-based MIL methods have shown better performance in many WSI analysis tasks\cite{transmil,HIPT,hag-mil}. However, the square complexity caused by the nonlinear mechanism of self-attention consumes more memory during training and inference, resulting in increased latency and reduced speed, which is not conducive to the actual deployment of algorithms in clinical scenarios.

To alleviate the above challenges, in this paper, we proposed a retentive multiple instance learning neural network called RetMIL, which introduces a retention mechanism to replace nonlinear self-attention, and effectively integrates subsequence information of WSI to obtain global representation with local features by building a hierarchical structure. We conduct experiments on public CAMELYON and BRACS datasets, as well as LUNG dataset for public data training and internal data testing. Results demonstrate that our proposed RetMIL achieves lower memory cost and higher throughput while exhibiting competitive performance.

\section{Methodology}

\begin{figure}[h]
  \centering
  \includegraphics[width=1\linewidth]{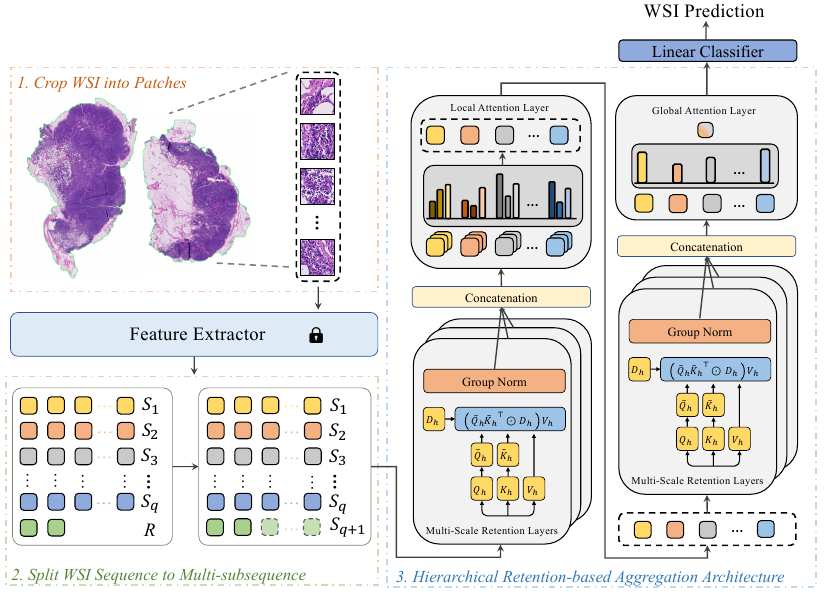}
   \caption{Overall framework of RetMIL}
   \label{fig:main_fig}
\end{figure}

In this section, we introduce the methodology of RetMIL. First, WSI is processed into a sequence form. Then local subsequences and the global sequence are sequentially updated and aggregated through retention and attention pooling. Finally, the prediction score of WSI is obtained through the classification head. Fig \ref{fig:main_fig} shows the overall framework of RetMIL.

\subsection{From WSI to Sequence}
We preprocess a WSI in four steps. First, we use the OTSU algorithm \cite{otsu} to segment the WSI foreground, and then use the sliding window operation to crop patches under a fixed magnification. Secondly, ViT-S/16 \cite{vit}, which is pretrained based on DINO \cite{DINO} on large-scale WSIs \cite{kang2023benchmarking}, is used as a feature extractor to encode each patch into a high-dimensional feature embedding $x_i \in \mathbb{R}^{d \times 1}$. Next, we form all $x_i$ into a sequence $X=\{x_1, \dots, x_N\}$, and split it into multiple subsequences $\{S_1,\dots,S_{q}, R\}$. Specifically, let $N=ql+r$, where $l$ is the length of each subsequence $S_j = \{x_{(j-1)l+1}, \dots, x_{jl}\}$, $j=1,2,\dots,q$ and $r=|R|$. Finally, to ensure that all subsequences have the same length and facilitate parallel calculation, we extend $R$ to $S_{q+1}=Concat(R, X_{l-r})$, where $|X_{l-r}|=l-r$, and there are three situations for $X_{l-r}$ as follows.
\begin{itemize}
    \item If $r = 0$, then $X_{l-r}=\varnothing$. 
    \item If $0 < r < l/2$, let $l-r=ar+b$ and $A=\{x_{ql+1}, \dots, x_{ql+r}\}$, then $X_{l-r} =\{\underbrace{{A,\dots,A}}_{\text{$a$ in total}},{x_{ql+1}}, \dots, x_{ql+b} \}  $.
    \item If $r \ge l/2$, then $X_{l-r}=\{x_{ql+1},\dots,x_{ql+(l-r)}\}$.
\end{itemize}

The purpose of processing $R$ in this way is to make each $x_i$ exist in only one subsequence, ensuring that the mapping between feature embeddings and subsequences is satisfied.

\subsection{Retention Mechanism}

Inspired by applications of the retentive network in large language models \cite{retnet}, RetMIL updates and aggregates sequence tokens through retention mechanisms. Given the matrix form $S \in \mathbb{R}^{|S| \times d}$ of an input sequence, we first use three linear layers to project it into different feature spaces:
\begin{equation}
    Q = XW_Q,K= XW_K,V = XW_V,
\end{equation}
where $W_Q$, $W_K$, and $W_V$ are learnable transformation matrices respectively. Next, we split $Q, K$ and $V$ into multiple heads $\{ Q_h \}$, $\{K_h \}$ and $\{ V_h \}$ and perform rotational position encoding \cite{su2024roformer} on each $Q_h$ and $K_h$ to obtain $\tilde{Q}_h$ and $\tilde{K}_h$. Then we use the retention layer for processing, which is expressed as follows:
\begin{equation}
    Retention(h,X)=(\tilde{Q}_h \tilde{K}_h^{\top} \odot D_h) V_h,
\end{equation}
where $D_h$ is a relative distance decay matrix, and each element $D_{h,nm}$ is expressed as:
\begin{equation}
    D_{h,nm} = \left\{\begin{matrix}
 \gamma^{n-m},  & n \ge m\\
 0,             & n < m
\end{matrix}\right.
\end{equation}
Finally, we use GroupNorm \cite{wu2018group} and swish gate \cite{hg16,rzl17} to normalize the output, and concatenate all retention head. The above mapping relationship can provide batch-level parallel calculation. When the input batch is $B$, we denote the entire update operation as $MSR(B;S)$.

\subsection{Hierarchical Retentive Aggregation Architecture}

For any subsequence matrix $S_i$,$i \in 1,\dots,q+1$ in WSI, $MSR(1;S_i)$ represents the result of $S_i$ after passing through the retention mechanism. Our goal is to update all subsequences in parallel, which is expressed as follows:

\begin{align}
 (F_1,\dots ,F_{q+1}) =& (MSR(1;S_1), \dots, MSR(1;S_{q+1}))\\ \notag
  =& MSR(q+1; (S_1, \dots, S_{q+1})),
\end{align}
where $F_i \in \mathbb{R}^{l \times d}$ represents the output embedding of subsequence $S_i$. Next, we use the attention pooling layer to aggregate the element features of each subsequence, which is expressed as:

\begin{equation}
    F_{local,i} = \sum_{k=1}^{l} \alpha_{i,k} F_{i,k},  
\end{equation}
where $F_{i,k}$ represents the $k$th element of $F_i$, $F_{local,i} \in \mathbb{R}^{d \times 1}$ represents the feature embedding of subsequence $S_i$. $\alpha_k$ is calculated through a nonlinear gating mechanism:

\begin{equation}
    \alpha_{i,k} = \frac{\exp\{ \mathrm{\Gamma}_l \tanh(\mathrm{W}_l F_{i,k})  \odot \mathrm{sigm}(\mathrm{U}_l F_{i,k})   \}}{\sum_{t=1}^{l} \exp\{ \mathrm{\Gamma}_l \tanh(\mathrm{W}_l F_{i,t})  \odot \mathrm{sigm}(\mathrm{U}_l F_{i,t})   \}},
\end{equation}
where $\mathrm{\Gamma}_l \in \mathbb{R}^{1 \times M}$, $\mathrm{W}_l,\mathrm{U}_l \in \mathbb{R}^{M \times d}$ are learnable parameters, $\tanh(\cdot), \mathrm{sigm}(\cdot)$ are nonlinear activation functions based on tanh and sigmoid respectively.

Next, we convert the feature embeddings of all subsequences into the local WSI feature matrix $F_{local} = (F_{local,1}, \dots, F_{local,q+1})^{\top} \in \mathbb{R}^{(q+1) \times d}$, and utilize the retention mechanism to update:

\begin{equation}
    G = MSR(1;F_{local}),
\end{equation}
where $G \in \mathbb{R}^{(q+1) \times d}$. Then attention pooling is used again to aggregate the $(q+1)$ dimension:

\begin{equation}
    F_{global} = \sum_{p=1}^{q+1} \beta_{p} G_{p},
\end{equation}
where $G_{p}$ represents the $p$th row element of $G$, and $\beta_p$ represents as follows:

\begin{equation}
    \beta_p = \frac{\exp\{ \mathrm{\Gamma}_{global} \tanh(\mathrm{W}_{global} G_p)  \odot \mathrm{sigm}(\mathrm{U}_{global} G_p)   \}}{\sum_{t=1}^{q+1} \exp\{ \mathrm{\Gamma}_{global} \tanh(\mathrm{W}_{global} G_t)  \odot \mathrm{sigm}(\mathrm{U}_{global} G_t) \}},
\end{equation}
where $\mathrm{\Gamma}_{global} \in \mathbb{R}^{1 \times M}$, $\mathrm{W}_{global},\mathrm{U}_{global} \in \mathbb{R}^{M \times d}$ are learnable parameters. For the WSI classification task, $F_{global}$ is passed through a linear classifier to obtain the prediction score. The entire RetMIL is trained using the cross-entropy function as the objective loss.

\section{Experiment}
\subsection{Datasets}
\textbf{CAMELYON: }The CAMELYON dataset focuses on the binary classification task of lymph node metastases in breast cancer. It includes 399 WSIs from CAMELYON16 \cite{cam16} and 500 WSIs from CAMELYON17 \cite{cam17}. We use all data of CAMELYON16 to conduct four-fold cross-validation experiments, and choose the CAMELYON17 training set as our testing dataset.

\noindent \textbf{BRACS:} The BRACS dataset \cite{bracs} focuses on multi-classification tasks aimed at subtype analysis of breast cancer. We conduct experiments based on the official classification. The dataset comprises 395 training samples, 65 validation samples, and 87 test samples. We use four different sets of model initialization parameters for training and testing.

\noindent \textbf{LUNG:} The LUNG dataset is a binary classification task focusing on non-small cell lung cancer subtypes. The training set is collected from TCGA repository \cite{tcga}, containing 541 WSIs of lung adenocarcinoma (LUAD) and 458 lung squamous cell carcinoma (LUSC). The test set is from the cooperative hospital, comprising 105 LUAD and 65 LUSC WSIs. We conduct four-fold cross-validation experiments on the training set and perform inference on the test set.

\subsection{Experiment Setup and Evaluation Metrics}

During the preprocessing stage, all WSIs are cropped into $224\times224$ patches at $20\times$ magnification. The length of each subsequence is set to 512. The entire experiment is conducted on one NVIDIA RTX 4090, with 100 epochs, utilizing early stopping with 15 rounds. The batch size is set as 1, and the learning rate is 1e-4, with a weight decay of 1e-5 for the Adam optimizer. All other baseline methods adopt the same experimental settings. We record the Balanced Accuracy(B-Acc) and Weighted F1-score as evaluation metrics to comprehensively evaluate the performance.

\subsection{Result}

\subsubsection{Performance Evaluation against SOTA:}


Table \ref{sota1} displays the performance of our proposed RetMIL, and we compare it with the following six state-of-the-art methods: For attention-based MIL: ABMIL \cite{abmil}, DSMIL \cite{dsmil}, CLAM-MB \cite{clam}. For Transformer-based MIL: TransMIL\cite{transmil}, HIPT \cite{HIPT} and HAG-MIL \cite{hag-mil}.
In the CAMELYON dataset, our RetMIL surpasses the second-ranked model TransMIL by 3.18\% and 3.43\% in F1-score and balanced accuracy. In the BRACS dataset, our model leads by 1.52\% and 0.86\% compared with the second-ranked CLAM-MB, while also achieving the minimum variance among all models. In the LUNG dataset, RetMIL outperforms by 0.13\%  in balanced accuracy. 

\begin{table}[h]
\caption{Mean and standard deviation of F1-score and Balanced accuracy (expressed in \%) between RetMIL and current powerful MIL method. The best is in \textbf{BOLD}, and the second best is indicated with \underline{underline}.}
\centering 
\setlength\tabcolsep{8pt}   
\resizebox{1\columnwidth}{!}{
\renewcommand{\arraystretch}{1.5}{
\begin{tabular}{lcccccc}
\hline
\multirow{2}{*}{\textbf{Methods}} & \multicolumn{2}{c}{\textbf{CAMELYON}} & \multicolumn{2}{c}{\textbf{BRACS}} & \multicolumn{2}{c}{\textbf{LUNG}} \\ \cline{2-7} 
 & \textbf{F1-score} & \textbf{B-Acc} & \textbf{F1-score} & \textbf{B-Acc} & \textbf{F1-score} & \textbf{B-Acc} \\ \hline
\textbf{ABMIL \cite{abmil}} & $81.27_{3.11}$ & \multicolumn{1}{c|}{$81.60_{2.30}$} & $64.11_{5.24}$ & \multicolumn{1}{c|}{$63.17_{4.39}$} & $88.68_{3.98}$ & $90.71_{3.26}$ \\
\textbf{CLAM-MB \cite{clam}} & $83.06_{4.59}$ & \multicolumn{1}{c|}{$83.37_{3.15}$} & \underline{$66.99_{4.02}$} & \multicolumn{1}{c|}{\underline{$66.15_{3.65}$}} & $87.67_{2.25}$ & $89.73_{1.76}$ \\
\textbf{DSMIL \cite{dsmil}} & $83.98_{1.79}$ & \multicolumn{1}{c|}{$83.77_{1.30}$} & $60.12_{4.52}$ & \multicolumn{1}{c|}{$59.22_{3.23}$} & $85.86_{9.15}$ & $86.57_{8.18}$ \\ 
\textbf{TransMIL \cite{transmil}} & \underline{$84.06_{8.19}$} & \multicolumn{1}{c|}{\underline{$84.10_{5.37}$}} & $62.83_{3.97}$ & \multicolumn{1}{c|}{$61.56_{3.53}$} & $\mathbf{91.75_{2.73}}$ & \underline{$91.43_{3.51}$} \\
\textbf{HIPT \cite{HIPT}} & $78.92_{8.11}$ & \multicolumn{1}{c|}{$80.17_{5.52}$} & $66.19_{8.97}$ & \multicolumn{1}{c|}{$65.73_{6.92}$} & $81.55_{6.37}$ & $84.85_{4.83}$ \\
\textbf{HAG-MIL \cite{hag-mil}} & $79.35_{5.71}$ & \multicolumn{1}{c|}{$80.59_{4.08}$} & $66.26_{4.52}$ & \multicolumn{1}{c|}{$64.76_{4.80}$} & $85.47_{4.42}$ & $87.61_{3.57}$ \\ \hline
\textbf{RetMIL (Ours)} & $\mathbf{87.24_{4.22}}$ & \multicolumn{1}{c|}{$\mathbf{87.53_{3.92}}$} & $\mathbf{68.51_{0.54}}$ & \multicolumn{1}{c|}{$\mathbf{67.01_{0.71}}$} & \underline{${91.51_{2.64}}$} & $\mathbf{91.56_{2.77}}$ \\ \hline
\end{tabular}}}

\label{sota1}
\end{table}

\begin{figure}[h]
  \centering
  \includegraphics[width=0.95\linewidth]{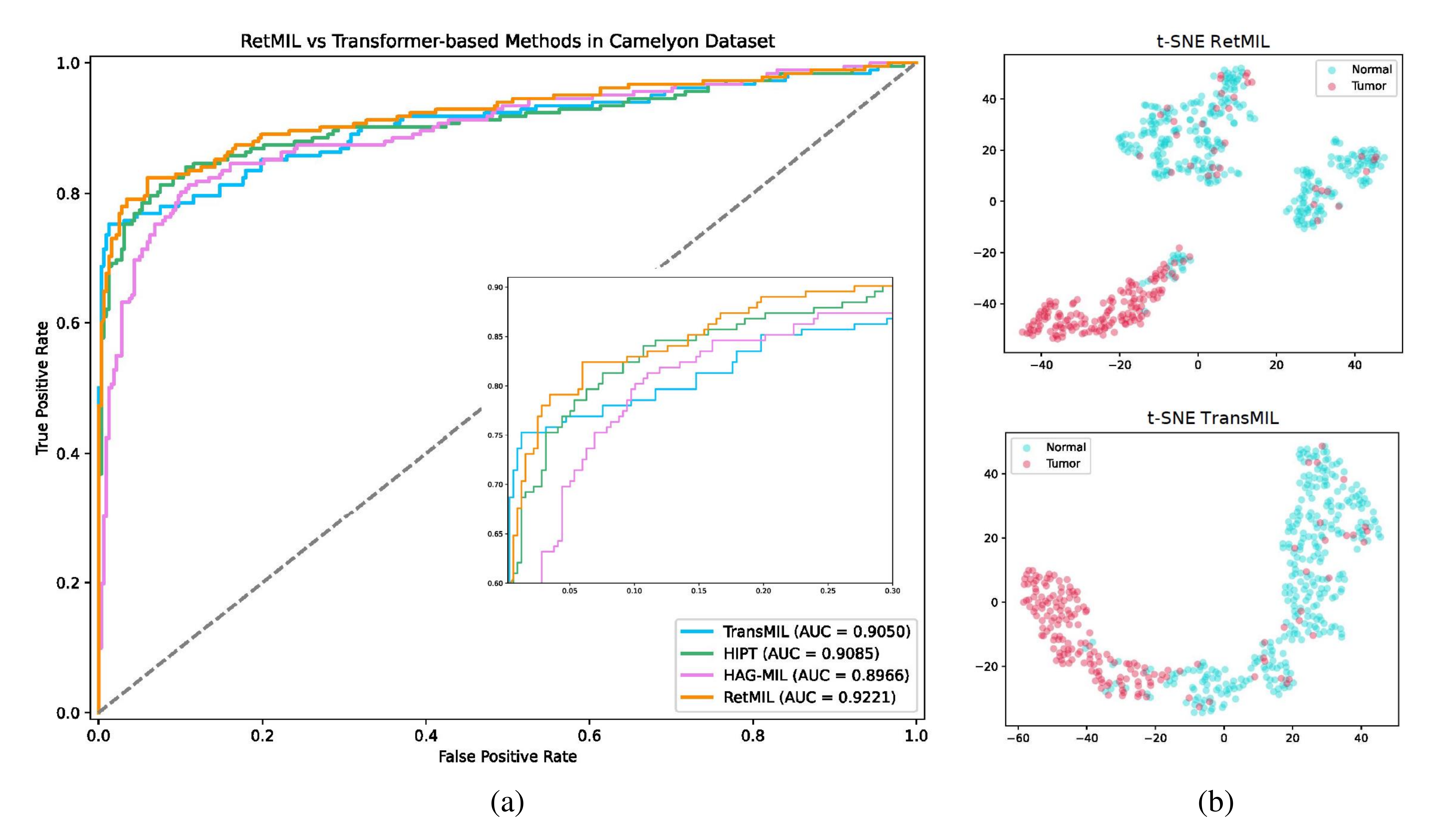}
   \caption{(a) ROC curves and corresponding area under the curve(AUC) values for RetMIL and Transformer-based models (b)Visual analysis of feature dimensionality reduction between RetMIL and TransMIL.}
   \label{fig:auc_tsne}
\end{figure}

We also compared RetMIL with Transformer-based models on the CAMELYON dataset using AUC, which are shown in Fig \ref{fig:auc_tsne}.a. It can be observed that RetMIL achieves a 1.36\% improvement in AUC compared to Transformer-based models. Additionally, Fig \ref{fig:auc_tsne}.b demonstrates the results of feature representations. All feature embeddings are reduced to a two-dimensional vector through the t-SNE algorithm \cite{tsne}. Our observation reveals that RetMIL can better widen the gap between distinct categories while minimizing the separation among patches belonging to the same category compared with the TransMIL algorithm.

\subsubsection{Performance at different lengths of sequences:}

In Transformer-based MIL methods, the length of the WSI sequence represents the number of cropped patches. We analyze model performance under different sequence lengths, and the result is shown in Table \ref{table:sequ}. Regardless of the length of the WSI sequence, our proposed method always significantly outperforms Transformer-based methods, especially for ultra-long sequences (i.e., oversized WSI), which demonstrates the effectiveness of RetMIL in long sequence analysis.

\begin{table}[h]
\caption{Performance comparison of RetMIL and Transformer-based models at different sequence lengths.}
\resizebox{1.\columnwidth}{!}{
\renewcommand{\arraystretch}{1.5}{
\begin{tabular}{lcccccccc}
\hline
\multirow{3}{*}{\textbf{Methods}} & \multicolumn{8}{c}{\textbf{Patch Number}} \\ \cline{2-9} 
 & \multicolumn{2}{c}{\textbf{0-5000}} & \multicolumn{2}{c}{\textbf{5001-10000}} & \multicolumn{2}{c}{\textbf{10001-15000}} & \multicolumn{2}{c}{\textbf{15001-}} \\ \cline{2-9} 
 & \textbf{F1-score} & \textbf{B-Acc} & \textbf{F1-score} & \textbf{B-Acc} & \textbf{F1-score} & \textbf{B-Acc} & \textbf{F1-score} & \textbf{B-Acc} \\ \hline
\textbf{TransMIL \cite{transmil}} & $81.83_{8.33}$ & \multicolumn{1}{c|}{$81.69_{5.55}$} & $86.66_{8.29}$ & \multicolumn{1}{c|}{$86.51_{5.37}$} & $84.47_{10.46}$ & \multicolumn{1}{c|}{${84.78_{8.35}}$} & $79.29_{5.83}$ & $79.89_{4.95}$ \\
\textbf{HIPT \cite{HIPT}} & \multicolumn{1}{c}{$77.98_{6.89}$} & \multicolumn{1}{c|}{$77.63_{4.74}$} & \multicolumn{1}{c}{$83.86_{8.59}$} & \multicolumn{1}{c|}{$84.34_{5.95}$} & \multicolumn{1}{c}{$79.89_{4.93}$} & \multicolumn{1}{c|}{$81.43_{3.08}$} & \multicolumn{1}{c}{$73.57_{4.45}$} & \multicolumn{1}{c}{$74.17_{3.81}$} \\
\textbf{HAG-MIL \cite{hag-mil}} & $78.10_{5.74}$ & \multicolumn{1}{c|}{$78.18_{4.41}$} & $82.32_{5.81}$ & \multicolumn{1}{c|}{$83.39_{4.13}$} & $78.74_{6.20}$ & \multicolumn{1}{c|}{$80.72_{4.80}$} & $68.73_{7.15}$ & $71.14_{5.13}$ \\ \hline
\textbf{RetMIL (Ours)} & $\mathbf{86.67_{2.26}}$ & \multicolumn{1}{c|}{$\mathbf{83.99_{1.70}}$} & $\mathbf{89.76_{1.85}}$ & \multicolumn{1}{c|}{$\mathbf{88.19_{1.58}}$} & $\mathbf{88.59_{0.60}}$ & \multicolumn{1}{c|}{$\mathbf{87.64_{0.42}}$} & $\mathbf{82.63_{4.42}}$ & $\mathbf{82.50_{4.20}}$ \\ \hline
\end{tabular}}}
\label{table:sequ}
\end{table}

\subsubsection{Inference performance:}

\begin{figure}[h]
  \centering
  \includegraphics[width=1 \linewidth]{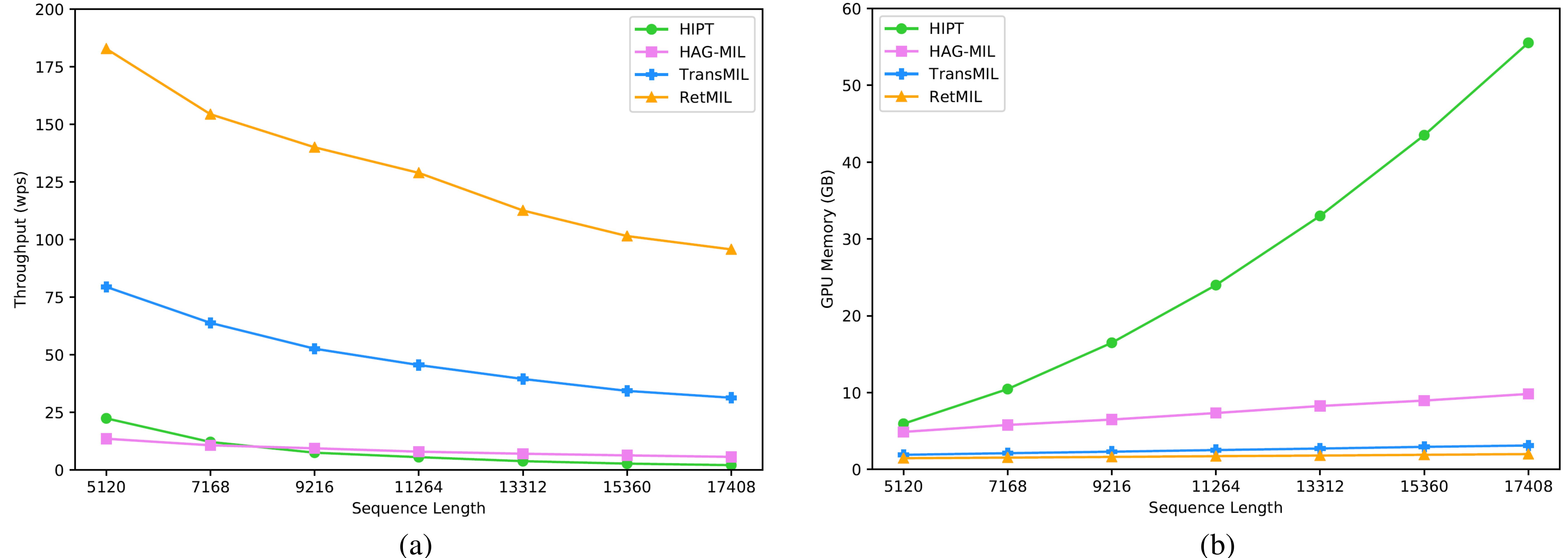}
   \caption{(a) Comparison of throughput between RetMIL and Transformer-based models at different sequence lengths. (b) Comparison of GPU memory consumption between RetMIL and Transformer-based models at different sequence lengths.}
   \label{fig:onecol}
\end{figure}

We also analyze the inference throughput and GPU memory usage under different sequence lengths with the Transformer-based models. As shown in Fig \ref{fig:onecol}.b, the GPU memory consumption of HIPT and HAG-MIL almost linearly increases with increasing sequence lengths, except for the lightweight-designed TransMIL model. However, our RetMIL maintains almost constant GPU memory consumption. Fig \ref{fig:onecol}.a shows that our retention model significantly improves model throughput. Even compared to the lightweight-designed TransMIL, our model maintains a nearly $1.5\times$ lead in throughput. 

\subsubsection{Visulization:}

\begin{figure}[h]
  \centering
  \includegraphics[width=1 \linewidth]{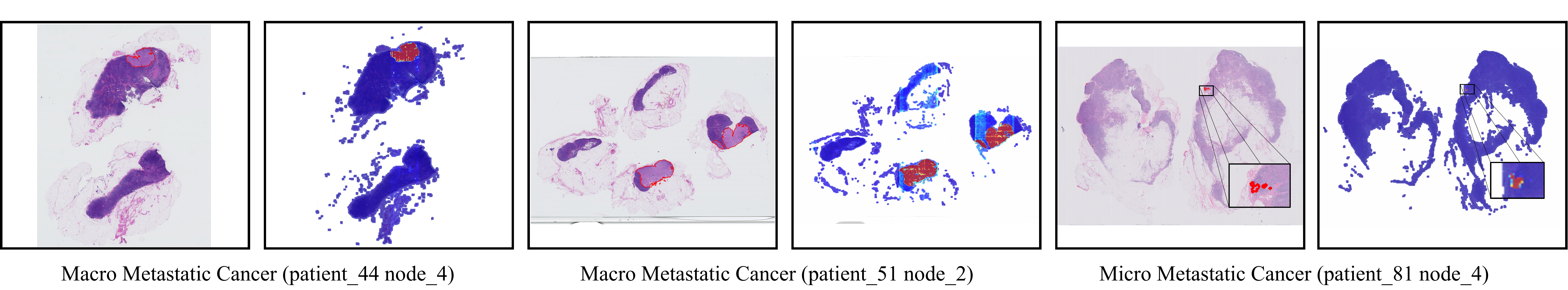}
   \caption{Heatmap visualization of WSI examples. In each pair of images, the left part displays the standard cancer regions outlined by pathologists (indicated by red contours), while the right side shows the heatmaps generated by our RetMIL.}
   \label{fig:visual}
\end{figure}

Fig \ref{fig:visual} presents heatmap visualization results of our RetMIL. We select two macro-metastatic cancer slides and one micro-metastasis cancer slide from the CAMELYON17 to analyze the attention area of our model. For the $k$th element in subsequence $i$, the attention score $score_{i,k}$ can be calculated as follow:
\begin{equation}
    s_{i,k} = \alpha_{i,k} \cdot \beta_{i},
\end{equation}
\noindent For both macro-metastatic and micro-metastatic cancer, our model can accurately and comprehensively pay attention to the cancer area marked by the pathologist, which demonstrates the great interpretability of our model.

\section{Conclusion}
In this paper, we propose a retentive multiple instance learning approach called RetMIL, which uses linear retention mechanisms to reduce the computational overhead while modeling the correlation between patches. In addition, the hierarchical retentive aggregation architecture is designed to update local subsequences and characterize the global WSI sequence comprehensively. We demonstrate the superiority of RetMIL through comparative experiments on three histopathology WSI datasets. At the same time, we also compared the inference performance with the Transformer-based methods, and the results show that our proposed RetMIL has lower computational consumption.


%
%

\bibliographystyle{splncs04}
\bibliography{ref}

\end{document}